\documentclass[runningheads]{llncs}
\usepackage[T1]{fontenc}
\usepackage{graphicx}
\usepackage{tikz}
\usetikzlibrary{shapes.geometric, arrows, positioning}

\usepackage{pgfplots}
\pgfplotsset{compat=1.17}
\usepackage{appendix}
\usepackage{booktabs}
\usepackage{subcaption}
\usepackage{array}
\usepackage{amsmath}
\usepackage{multirow}
\begin{document}
\title{USAM-Net: A U-Net-based Network for Improved Stereo Correspondence and Scene Depth Estimation using Features from a Pre-trained Image Segmentation network}
%
%
\author{Joseph Emmanuel DL Dayo\inst{1}\orcidID{0009-0007-6943-2003} \and \\
Prospero C. Naval Jr.\inst{2}\orcidID{0000-0001-7140-1707}}
%

%
\institute{University of the Philippines, Diliman, Quezon City, Philippines  \\
\email{jddayo@up.edu.ph}\\
\and
\email{pcnaval@up.edu.ph}}
\titlerunning{USAM-Net (Unified Segmentation Attention Mechanism Network)}
\maketitle              
\begin{abstract}
The increasing demand for high-accuracy depth estimation in autonomous driving and augmented reality applications necessitates advanced neural architectures capable of effectively leveraging multiple data modalities. In this context, we introduce the Unified Segmentation Attention Mechanism Network (USAM-Net), a novel convolutional neural network that integrates stereo image inputs with semantic segmentation maps and attention to enhance depth estimation performance. USAM-Net employs a dual-pathway architecture, which combines a pre-trained segmentation model (SAM) and a depth estimation model. The segmentation pathway preprocesses the stereo images to generate semantic masks, which are then concatenated with the stereo images as inputs to the depth estimation pathway. This integration allows the model to focus on important features such as object boundaries and surface textures which are crucial for accurate depth perception.  Empirical evaluation on the DrivingStereo dataset demonstrates that USAM-Net achieves superior performance metrics, including a Global Difference (GD) of 3.61\% and an End-Point Error (EPE) of 0.88, outperforming traditional models such as CFNet, SegStereo, and iResNet. These results underscore the effectiveness of integrating segmentation information into stereo depth estimation tasks, highlighting the potential of USAM-Net in applications demanding high-precision depth data.

\end{abstract}
\section{Introduction}

In computer vision, stereo matching and calculating for disparity are essential in order to properly compute the depth of the scene. Accurate depth calculation is useful for various tasks like autonomous driving, vision-based robot navigation, 3D scene reconstruction, among other things.

Estimating depth from stereo images requires a process known as stereo matching, where features along the epipolar lines are matched from the left to the right images in order to compute the disparity. However, the presence of featureless areas and occlusion can limit the effectiveness of such algorithms. This problem is exacerbated in  autonomous driving scenarios where streets are dominated by smooth and featureless surfaces like roads, flat vehicles, shadows and the sky.

Given the importance of depth estimation in various tasks like autonomous driving and motion planning, how do we improve the robustness and accuracy of deep learning methods by incorporating additional features and structures inside a neural network?

Many researchers have tried to solve the problem of matching disparity in featureless regions with varying degrees of success. Yang et al. \cite{yang2018segstereo} used an end-to-end model that employs segmentation maps to add semantic information to stereo images, with ResNet-50 as the backbone. For USAM-Net, a similar process is employed; however, we focus more on an existing pre-trained state-of-the-art segmentation model known as the Segment Anything Model (SAM) \cite{kirillov2023segany}. To offset the overhead due to SAM, we opted to use a ResNet-18-based model for efficiency.

Recent advancements in depth estimation techniques have leveraged the integration of semantic segmentation with traditional convolutional neural networks (CNNs) to enhance the accuracy and reliability of monocular depth predictions. A notable approach, as discussed by Cantrell et. al \cite{cantrell2020pracdepth}, employs serial U-Net architectures which modularly combine multiple U-Nets pre-trained on diverse tasks such as image segmentation and object detection to improve depth estimation. This method capitalizes on the robust feature extraction capabilities of U-Nets and integrates them in a serial configuration, allowing for the sequential processing of learned features across the network. This architecture has shown promising results in reducing hardware dependencies and computational costs associated with conventional depth estimation technologies like LIDAR and stereo vision systems, thereby presenting a cost-effective solution for autonomous vehicle navigation and intelligent transport systems. The integration of pre-trained networks enhances the system’s ability to interpret complex scenes, making it a significant step forward in the application of deep learning to real-world problems in depth estimation.

Building on the existing literature, a more recent study by Jan and Seo \cite{abdullah2023monodepth} introduces a novel approach to monocular depth estimation utilizing a Res-UNet architecture combined with a spatial attention model. This method addresses the challenge of accurate depth prediction from single images by enhancing feature extraction capabilities and focusing on boundary definitions without increasing the computational cost through additional parameters. The implementation of this attention mechanism within a convolutional neural network framework demonstrates significant improvements over previous state-of-the-art methods on the NYU Depth v2 dataset \cite{Silberman:ECCV12}, marking a substantial advancement in the field of depth estimation technologies. This development aligns with the growing trend towards more efficient and accurate depth sensing solutions in applications ranging from autonomous navigation to 3D mapping.

Huang et al. \cite{huang2022sdepth} introduced H-Net, an innovative approach to unsupervised stereo depth estimation leveraging epipolar geometry. H-Net employs a Siamese autoencoder architecture that enhances stereo matching by focusing on mutual epipolar attention. This mechanism effectively emphasizes feature correspondences along epipolar lines, crucially integrating semantic information to improve the reliability of the depth estimates. Extensive testing on the KITTI2015 and Cityscapes datasets demonstrated H-Net’s capability to close the performance gap with fully supervised methods, marking a significant advancement in the field of computer vision for autonomous systems and 3D mapping.

Finally, Oktay et. al \cite{attentionunet} introduced Attention U-Net, primarily designed for medical imaging, that utilizes attention blocks in between the U-Net skip connections and automatically learns to focus on target structures of varying shapes and sizes.

\section{Methodology}

In this paper, we propose USAM-Net, a self-attention-based neural network that seeks to provide state-of-the-art depth estimation using multiple stereo features as input. A set of images from the stereo camera and another set of images preprocessed using a pre-trained segmentation neural network will serve as features for the proposed neural network. For the segmentation neural network, we will use the Segment Anything model \cite{kirillov2023segany}. The Segment Anything model will then be used to generate segmentation images from the left stereo images of the DrivingStereo dataset \cite{yang2019drivingstereo}. These segmentation images will be used alongside the standard left and right images as additional input for training.

\subsection{USAM-Net}

The proposed model uses a U-Net based \cite{unetpaper} architecture, with an encoder network and a decoder network containing skip connections at various levels (Fig. \ref{UH-Net}). The input layer accepts stereo images as well as a an additional image for a segmentation map generated by the SAM model. We also propose adding an attention layer in between the encoder and decoder blocks though we will have variations that do not have this. For the purposes of predicting disparity for each pixel we use a sigmoid activation in the output layer and 2 additional convolutional layers after the encoder-decoder module. Another difference from the original U-Net architecture is the incorporation of batch normalization at each encoder and decoder layer. Padding and stride values were also adjusted to fit the target resolution of the dataset. Finally, we also added an attention layer (Fig. \ref{Self-Attention Layer}) in between the encoder and decoder block. Details of USAM-Net are provided in Table 1.

\subsubsection{Parameter Size and Performance} The USAM-Net base model is relatively lightweight and consists of 15.1M parameters with an inference speed of 2.18ms on an RTX 3090, by itself suitable for real-time applications. However, using the segmentation pathway would require taking into consideration SAM-Vit-b's 93.7M parameter count and 900ms inference speed.

\tikzstyle{layer} = [rectangle, rounded corners, minimum width=3cm, minimum height=1cm,text centered, draw=black, fill=blue!30]
\tikzstyle{arrow} = [thick,->,>=stealth]

\usetikzlibrary{shapes.geometric, arrows}

\tikzstyle{layer} = [rectangle, rounded corners, minimum width=4cm, minimum height=1cm, text centered, draw=black, fill=blue!30, align=center]
\tikzstyle{io} = [rectangle, rounded corners, minimum width=4cm, minimum height=1cm, text centered, draw=black, fill=green!30, align=center]
\tikzstyle{arrow} = [thick,->,>=stealth]
\tikzstyle{skip} = [thick,->,>=stealth, dashed]

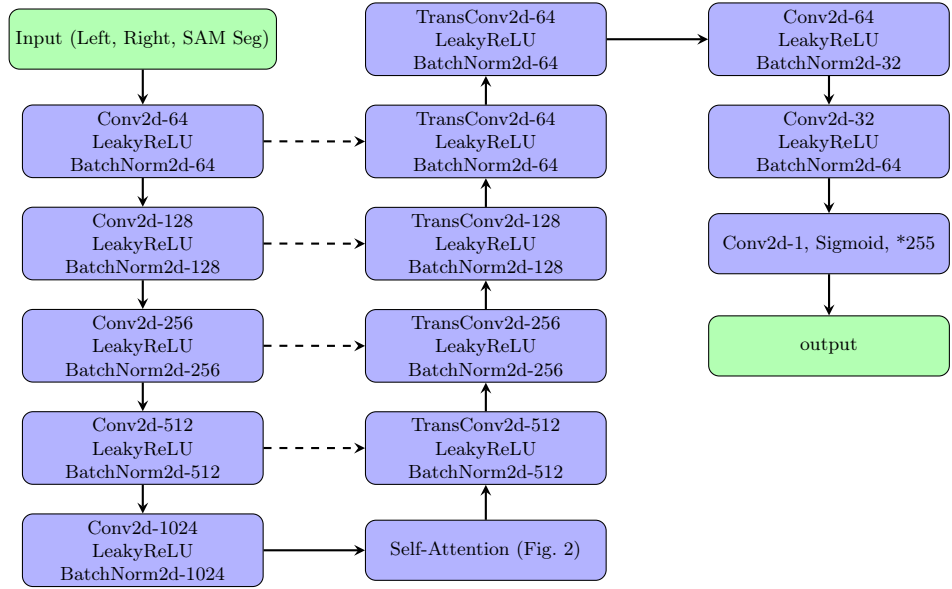
\begin{figure}[htbp]
    \centering
\begin{tikzpicture}[scale=0.8, node distance=1.7cm, transform shape]
    \node (input) [io] {Input (Left, Right, SAM Seg)};
    \node (down1) [layer, below of=input] {Conv2d-64 \\LeakyReLU \\ BatchNorm2d-64};
    \node (down2) [layer, below of=down1] {Conv2d-128\\LeakyReLU\\BatchNorm2d-128};
    \node (down3) [layer, below of=down2] {Conv2d-256 \\ LeakyReLU \\ BatchNorm2d-256};
    \node (down4) [layer, below of=down3] {Conv2d-512 \\ LeakyReLU \\ BatchNorm2d-512};
    \node (down5) [layer, below of=down4] {Conv2d-1024 \\ LeakyReLU \\ BatchNorm2d-1024};

    \node (sa1) [layer, right of=down5, xshift=4cm] {Self-Attention (Fig. \ref{Self-Attention Layer})};

    \node (up1) [layer, above of=sa1] {TransConv2d-512 \\ LeakyReLU \\ BatchNorm2d-512};
    \node (up2) [layer, above of=up1] {TransConv2d-256 \\ LeakyReLU \\ BatchNorm2d-256};
    \node (up3) [layer, above of=up2] {TransConv2d-128 \\ LeakyReLU \\ BatchNorm2d-128};
    \node (up4) [layer, above of=up3] {TransConv2d-64 \\ LeakyReLU \\ BatchNorm2d-64};
    \node (up5) [layer, above of=up4] {TransConv2d-64 \\ LeakyReLU \\ BatchNorm2d-64};

    \node (conv1) [layer, right of=up5, xshift=4cm] {Conv2d-64 \\ LeakyReLU \\ BatchNorm2d-32};

    \node (conv2) [layer, below of=conv1] {Conv2d-32 \\ LeakyReLU \\ BatchNorm2d-64};

    \node (conv3) [layer, below of=conv2] {Conv2d-1, Sigmoid, *255};

    \node (output) [io, below of=conv3] {output};

    \draw [arrow] (input) -- (down1);
    \draw [arrow] (down1) -- (down2);
    \draw [arrow] (down2) -- (down3);
    \draw [arrow] (down3) -- (down4);
    \draw [arrow] (down4) -- (down5);

    \draw [arrow] (down5) -- (sa1);
    \draw [arrow] (sa1) -- (up1);
    \draw [arrow] (up1) -- (up2);
    \draw [arrow] (up2) -- (up3);
    \draw [arrow] (up3) -- (up4);
    \draw [arrow] (up4) -- (up5);
    \draw [arrow] (up5) -- (conv1);
    \draw [arrow] (conv1) -- (conv2);
    \draw [arrow] (conv2) -- (conv3);
    \draw [arrow] (conv3) -- (output);

    \draw [skip] (down1.east) -- (up4.west);
    \draw [skip] (down2.east) -- (up3.west);
    \draw [skip] (down3.east) -- (up2.west);
    \draw [skip] (down4.east) -- (up1.west);

\end{tikzpicture}
    \caption{Basic Architecture of USAM-Net}
    \label{UH-Net}
\end{figure}
\

\tikzstyle{startstop} = [rectangle, rounded corners, minimum width=2cm, minimum height=0.5cm,text centered, draw=black, fill=red!30, align=center]
\tikzstyle{process} = [rectangle, minimum width=1cm, minimum height=0.5cm, text centered, draw=black, fill=orange!30, align=center]
\tikzstyle{arrow} = [thick,->,>=stealth]
\begin{figure}[htbp]
    \centering
\begin{tikzpicture}[scale=1.0, node distance=0.8cm, transform shape]
\centering
\node (input) [startstop] {Input $x$};
\node (query) [process, below of=input, xshift=-2cm] {Query};
\node (key) [process, below of=input] {Key};
\node (value) [process, below of=input, xshift=2cm] {Value};
\node (matmul1) [process, below of=key] {Matrix Mult.};
\node (softmax) [process, below of=matmul1] {Softmax};
\node (matmul2) [process, below of=softmax] {Matrix Mult.};
\node (reshape) [process, below of=matmul2] {Reshape};
\node (skip) [process, below of=reshape, xshift=4cm] {Skip Connection};
\node (output) [startstop, below of=skip] {Output};

\draw [arrow] (input) -| (query);
\draw [arrow] (input) -- (key);
\draw [arrow] (input) -| (value);
\draw [arrow] (query) |- (matmul1);
\draw [arrow] (key) -- (matmul1);
\draw [arrow] (matmul1) -- (softmax);
\draw [arrow] (softmax) -- (matmul2);
\draw [arrow] (value) |- (matmul2);
\draw [arrow] (matmul2) -- (reshape);
\draw [arrow] (reshape) -- (skip);
\draw [arrow] (input) -| (skip);
\draw [arrow] (skip) -- (output);

\end{tikzpicture}
\caption{The Self-Attention Layer}
\label{Self-Attention Layer}
\end{figure}
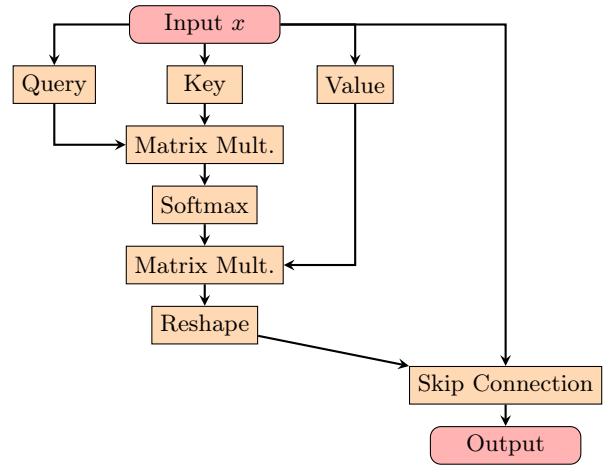

\begin{table}[ht!]
\centering
\caption{Detailed Architecture of USAM-Net}
{\footnotesize 
\renewcommand{\arraystretch}{1} 
\setlength{\tabcolsep}{5pt} 
\begin{tabular}{|>{\raggedright\arraybackslash}p{3cm}|>{\raggedright\arraybackslash}p{10cm}|}
\hline
\textbf{Layer (type)} & \textbf{Configuration} \\
\hline
Input Layer & Input image tensor (Left, Right, Segment mask from SAM for Left) \\
\hline
Conv2d-1 & 9 channels $\rightarrow$ 64 channels, 3x3 kernel, stride 2, padding 1 \\
LeakyReLU-1 & Negative slope 0.01 \\
BatchNorm2d-1 & 64 features \\
\hline
Conv2d-2 & 64 channels $\rightarrow$ 128 channels, 3x3 kernel, stride 2, padding 1 \\
LeakyReLU-2 & Negative slope 0.01 \\
BatchNorm2d-2 & 128 features \\
\hline
Conv2d-3 & 128 channels $\rightarrow$ 256 channels, 3x3 kernel, stride 2, padding 1 \\
LeakyReLU-3 & Negative slope 0.01 \\
BatchNorm2d-3 & 256 features \\
\hline
Conv2d-4 & 256 channels $\rightarrow$ 512 channels, 3x3 kernel, stride 2, padding 1 \\
LeakyReLU-4 & Negative slope 0.01 \\
BatchNorm2d-4 & 512 features \\
\hline
Conv2d-5 & 512 channels $\rightarrow$ 1024 channels, 3x3 kernel, stride 2, padding 1 \\
LeakyReLU-5 & Negative slope 0.01 \\
BatchNorm2d-5 & 1024 features \\
\hline
Self Attention  \\
\hline
ConvTranspose2d-1 & 1024 channels $\rightarrow$ 512 channels, 3x3 kernel, stride 2, pad 1, out pad 1 \\
LeakyReLU-6 & Negative slope 0.01 \\
BatchNorm2d-6 & 512 features \\
\hline
ConvTranspose2d-2 & 512 channels $\rightarrow$ 256 channels, 4x4 kernel, stride 2, pad 1, out pad 1 \\
LeakyReLU-7 & Negative slope 0.01 \\
BatchNorm2d-7 & 256 features \\
\hline
ConvTranspose2d-3 & 256 channels $\rightarrow$ 128 channels, 4x4 kernel, stride 2, pad 1, out padg 1 \\
LeakyReLU-8 & Negative slope 0.01 \\
BatchNorm2d-8 & 128 features \\
\hline
ConvTranspose2d-4 & 128 channels $\rightarrow$ 64 channels, 4x4 kernel, stride 2, pad 1, out pad 1 \\
LeakyReLU-9 & Negative slope 0.01 \\
BatchNorm2d-9 & 64 features \\
\hline
ConvTranspose2d-5 & 64 channels $\rightarrow$ 32 channels, 4x3 kernel, stride 2, pad 1, out pad 1 \\
LeakyReLU-10 & Negative slope 0.01 \\
BatchNorm2d-10 & 32 features \\
\hline
Final Conv Layers & Conv2d: 32 $\rightarrow$ 64 (3x3, stride 1, padding 1) $\rightarrow$ LeakyReLU \\
 & Conv2d: 64 $\rightarrow$ 128 (3x3, stride 1, padding 1) $\rightarrow$ LeakyReLU \\
 & Conv2d: 128 $\rightarrow$ 1 (1x1, stride 1, padding 0) \\
\hline
Output Layer & Sigmoid $\rightarrow$ Output disparity map scaled to 0-255 \\
\hline
\end{tabular}
}
\end{table}

\subsubsection{Overview of the DrivingStereo Dataset}

The DrivingStereo dataset consists of 174,437 training image stereo pairs including corresponding disparity maps. The test set, on the other hand, contains 7,751 stereo images and disparity maps. A combination of stereo cameras with LiDAR is used to generate depth and disparity information. The training and test stereo images have already been rectified and calibrated for the camera pairs. Both the depth map and disparity map are provided. However, we only use the disparity map during training

\begin{figure}[htbp]
    \centering
    \begin{subfigure}[b]{0.65\textwidth}
        \centering
        \includegraphics[width=0.6\textwidth]{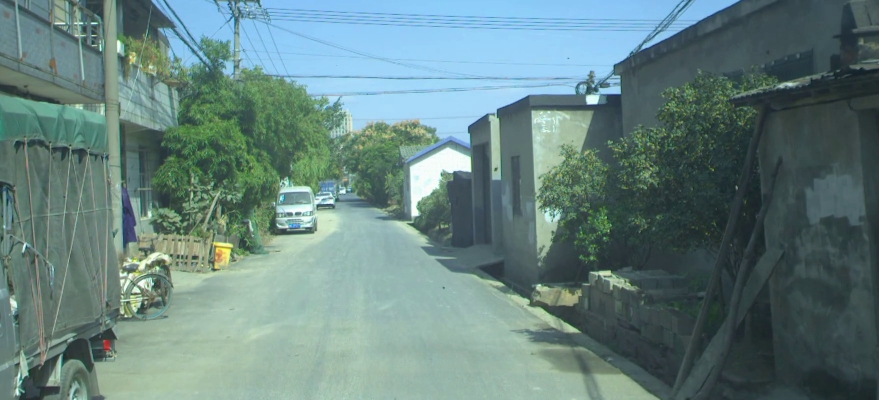}
        \caption{Left Image}
        \label{fig:left_image}
    \end{subfigure}
    \hfill
    \begin{subfigure}[b]{0.65\textwidth}
        \centering
        \includegraphics[width=0.6\textwidth]{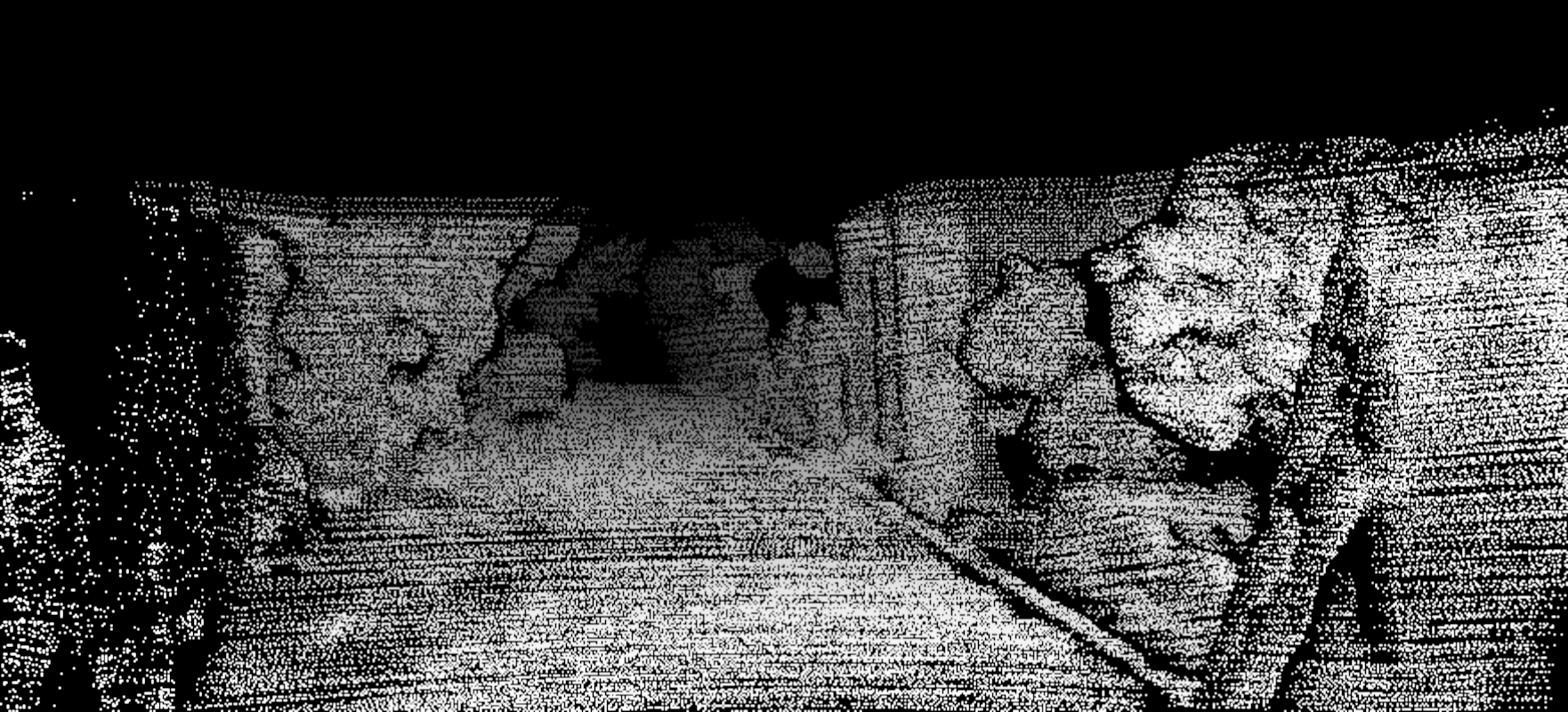}
        \caption{Disparity}
        \label{fig:disparity}
    \end{subfigure}
    \caption{Driving Stereo Dataset}
    \label{fig:side_by_side}
\end{figure}

Due to the sparsity of the disparity map with no ground truth on the upper part as well as even on some close objects, we mask the empty pixels during training.

\subsection{KITTI 2015 Dataset} The KITTI 2015 \cite{Menze2015CVPR} is a popular stereo benchmark developed by the Karlsruhe Institute of Technology
and Toyota Technological Institute that consists of 200 training scenes and 200 test scenes. While having scenes similar to the DrivingStereo dataset, it has a significantly smaller amount of samples. However, being an older and more popular dataset it is more widely used than the DrivingStereo dataset.

\subsection{Evaluation}

To evaluate the performance of our proposed network, we compare the performance against existing methodologies on the DrivingStereo dataset \cite{yang2019drivingstereo}. We will also test on variants of the USAM-Net model that includes: (1) Baseline (no Segmentation, no self-attention), (2) Baseline-2 (no Segmentation, with Self-Attention), (3) segmentation-only, and (4) with both segmentation and Self-Attention. The results of the four models are then  compared in order to gain additional insights on the architecture as well as for future studies.

\subsection{Metrics}

In order to compare the performance of our model with other studies on the DrivingStereo \cite{yang2019drivingstereo} dataset, we will adopt the EPE, D1, GD (Global Difference) and ARD (Absolute Relative Difference) metrics. Furthermore, as noted in the DrivingStereo paper \cite{yang2019drivingstereo}, the ARD and GD metrics is more appropriate for driving scenarios since it allows for equal treatment of near, middle and far objects, whereas the metrics like EPE (End Point Error) tend to over-emphasize nearer objects where the disparity is greatest.
\subsubsection*{D1 Error}

The D1 metric is a standard evaluation measure used in stereo matching tasks, especially in benchmarks like KITTI. It quantifies the percentage of pixels where the disparity estimation error exceeds a predefined threshold, focusing on significant errors to assess the robustness of stereo matching algorithms.\\
\\
The D1 error is defined as:

\begin{equation}
\text{D1 error} = \frac{\displaystyle \sum_{i=1}^{N} \delta\left( \left| d_i - d_i^{\text{gt}} \right| > \max\left( \tau_{\text{abs}}, \tau_{\text{rel}} \times d_i^{\text{gt}} \right) \right)}{N} \times 100\%
\end{equation}

where:

\begin{itemize}
    \item \( d_i \) is the estimated disparity at pixel \( i \).
    \item \( d_i^{\text{gt}} \) is the ground truth disparity at pixel \( i \).
    \item \( N \) is the total number of valid pixels in the ground truth disparity map.
    \item \( \delta(\cdot) \) is the indicator function, equal to 1 if the condition is true and 0 otherwise.
    \item \( \tau_{\text{abs}} \) is the absolute error threshold (typically 3 pixels).
    \item \( \tau_{\text{rel}} \) is the relative error threshold (typically 5\%).
\end{itemize}

A pixel is considered to have a disparity error if:

\[
\left| d_i - d_i^{\text{gt}} \right| > \max\left( \tau_{\text{abs}}, \tau_{\text{rel}} \times d_i^{\text{gt}} \right)
\]

The thresholds for this experiment are currently set to \( \tau_{\text{abs}} = 3 \) pixels and \( \tau_{\text{rel}} = 0.05 \). Therefore, the D1 metric calculates the percentage of pixels where the disparity error exceeds either 3 pixels or 5\% of the ground truth disparity.

\subsubsection{End Point Error}
The EPE measures the absolute difference between the predicted and ground truth disparity map values across all pixels in the image. Specifically, EPE is defined as the average Euclidean distance between the estimated disparity map and the ground truth disparity map. \\
\\
EPE can be expressed as follows:

\[
\text{EPE} = \frac{1}{N} \sum_{i=1}^{N} \left| d_i^{\text{pred}} - d_i^{\text{gt}} \right|
\]

where:
\begin{itemize}
    \item \( N \) is the total number of pixels.
    \item \( d_i^{\text{pred}} \) is the predicted disparity value at pixel \( i \).
    \item \( d_i^{\text{gt}} \) is the ground truth disparity value at pixel \( i \).
\end{itemize}

\subsubsection{ARD and GD}

The ARD is computed as follows:

\begin{equation}
\text{ARD}_k = \frac{1}{N_{R_k}} \sum_{d_g \in R_k} \frac{|d_p - d_g|}{d_g},
\end{equation}

where \( N_{R_k} \) is the number of valid pixels in \( R_k \). Each $k$ is an interval in the depth that we would like to measure. Disparity information will be obtained directly from the disparity map data that is part of the dataset. It is worth emphasizing that, due to the nature of the dataset, areas with no LIDAR coverage are set to zero pixel values. These areas will not be considered valid pixels and are excluded from the computation.

\paragraph{Hyperparameters:}
The following parameters were used for ARD and GD calculations, consistent with the DrivingStereo paper:
\begin{itemize}
    \item \textbf{Minimum depth (\( \text{min\_depth} \))}: 0
    \item \textbf{Maximum depth (\( \text{max\_depth} \))}: 80
    \item \textbf{Measuring range (\( r \))}: \(\pm 4\)
    \item \textbf{Sampling interval (\( \text{interval} \))}: 8
\end{itemize}

This results in 10 buckets, which start at 8, 16, 24,... 80. A range of 4 results in \(8\pm4\ or\ [4-12]\) for the first bucket, \(16\pm4\ or\ [23-20]\) and so on.

We also compute for the GD which is the overall summation of the ARDs:

\begin{equation}
\text{GD} = \frac{1}{K} \sum_{k \in K} \text{ARD}_k.
\end{equation}

\subsection{Masking the Sky and Top Portions}
\label{sec:data_preprocess}

The driving stereo dataset \cite{yang2019drivingstereo} is a dataset that consists of 174,437 stereo, disparity, and depth images. During analysis of the provided disparity and depth maps, we noticed that there were no valid pixels at the top quarter of the image, mostly due to the fact that LIDAR coverage is limited. Also, even those at the lower parts of the image, not all values of the depth and disparity maps are present, with some close objects not having any depth information. Initially, we planned to simply ignore the top part of the image (mask the predicted output so it is not included during backpropagation) but this resulted in artifacts showing up in the predicted disparity maps in the top parts. 

In an attempt to address this the following data pre-processing steps were performed: (1) computation of a custom mask for the upper portion of the image; (2) data augmentation that will randomly replace the top part of the image with either the middle or the bottom part. The same operation is performed on the disparity map as well.

For the custom mask, the SAM model is used to automatically create a segment for the sky portion of the image (the area where there is mostly no LIDAR data) and carve out a mask where we can assume a disparity of 0 for those. It should be noted that the GD and ARD metrics only consider valid pixels hence these data augmentation pre-processing will not play a role. However, this will greatly improve the qualitative properties of the final disparity map.

\begin{figure}[h]
    \centering
    \begin{subfigure}[b]{0.32\textwidth}
        \centering
        \includegraphics[width=\textwidth]{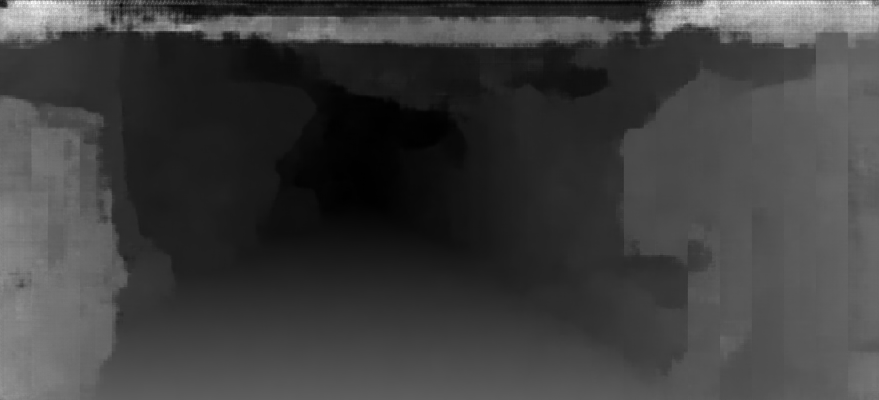}
        \caption{w/o Sky Masking}
    \end{subfigure}\hfill
    \begin{subfigure}[b]{0.32\textwidth}
        \centering
        \includegraphics[width=\textwidth]{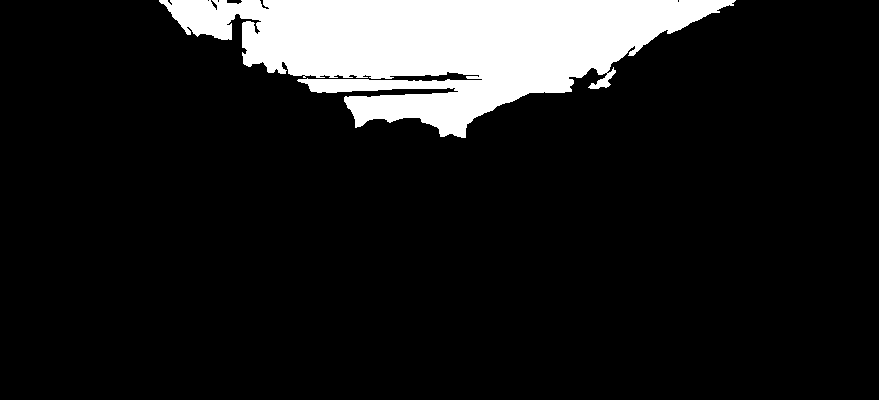}
        \caption{Sky Mask}
    \end{subfigure}\hfill
    \begin{subfigure}[b]{0.32\textwidth}
        \centering
        \includegraphics[width=\textwidth]{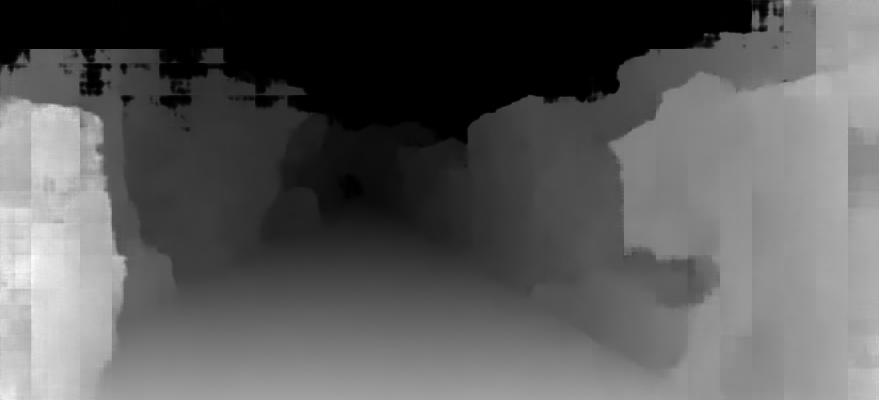}
        \caption{trained with sky masks}
    \end{subfigure}
    \caption{Comparison with masking the sky during training. Notice the artifacts on the top portion when there is no masking.}
\end{figure}

While this improves the qualitative output of the disparity, it was found out that this significantly reduces the accuracy on the test dataset, hence more investigation is needed on this approach.

\subsection{Experimental Setup}

For each of the models, we run 30 epochs of training using the Adam optimizer with a learning rate of 0.001 and a decay rate of 0.9 per epoch. For data augmentation, we add color jitter, which randomly varies the saturation, hue, contrast, and brightness by 10\%. Afterwards, the stereo image pairs in the dataset are normalized per channel using the specified mean and standard deviation values: mean = [0.50625, 0.52283, 0.41453], std = [0.21669, 0.19807, 0.18691]. To mitigate the effects of outliers in the disparity map, we use the smooth-L1~\cite{victor2021survey} loss function. Additionally, a masking function based on the training disparity is applied to the predicted disparity to consider only non-zero pixels.

To speed up training, we pre-processed all the left training and test images using the "base" SAM \textbf{ViT-B} model. Three Nvidia A100 GPUs were used in the training process, amounting to a total of 56 GPU hours. We then compare the ARD values at various depths for the three models. To ensure consistency and facilitate the experiments, we leverage a forked version of the OpenStereo~\cite{guo2023openstereo}\cite{githubGitHubJedldOpenStereo} framework for collecting and evaluating the results. Finally, we use the half-resolution (879×400) images from the DrivingStereo dataset.

\section{Results and Discussion}

We describe in Table \ref{performance_scores} the performance scores for all four model types:

\begin{table}[ht]
\centering
\caption{Performance Scores of Machine Learning Models on the DrivingStereo Dataset}
\label{my-label}
\begin{tabular}{@{}lcccc@{}}
\toprule
Model Type & EPE & D1 Error & GD & L1-loss\\ \midrule
\textbf{U-Net baseline (Ours)}& 0.964 & 2.53\% & 4.17\% & 0.12\\
\textbf{USAM-Net (Attention) (Ours)} & \textbf{0.889} & 1.94\% & 3.8\% & 0.10 \\
\textbf{USAM-Net (Segmentation) (Ours)}& 0.924 & 2.65\% & 3.68\% & 0.115 \\
\textbf{USAM-Net (Segmentation+Attention) (Ours)}& \textbf{0.88} & 2.26\% & \textbf{3.61}\% &0.116 \\
CFNet \cite{cfnet} & 0.98 & 1.46\% & -  & - \\
SegStereo \cite{yang2018segstereo} & 1.32 & 5.89\% & 4.78\% & - \\
EdgeStereo \cite{songxiaoedgestereo} & 1.19 & 3.47\% & 4.17\% & -\\
iResNet \cite{iresnet} & 1.24 & 4.27\% & 4.23\% & -\\ 
StereoBase \cite{guo2023openstereo} & 1.15 & 2.19\% & -  & - \\
IGEV-Stereo \cite{xu2023iterativegeometryencodingvolume}\cite{guo2023openstereo} & 1.06 & 1.50\% & - & - \\ 
\bottomrule
\end{tabular}
\label{performance_scores}
\end{table}

Based on the scores, we can see that our method (USAM-Net) results in the best GD and EPE scores  compared to other methods. Even the baseline U-Net model is equivalent to the EdgeStereo \cite{songxiaoedgestereo} model on the GD metric. Compared to the U-Net Baseline (same model but only with stereo input) we see that the segmentation information provided by SAM increases the accuracy of the USAM-Net models. It is interesting to note that while the Self-Attention layer is not having much of an effect when segmentation information is used, it certainly helps, as compared to the baseline, where segmentation is not present. This model variant can prove to be an alternative when the increased compute required for using the Segment Anything model is not computationally feasible for a use case.

Here we present the ARD curves of the various model variants. These plots indicate the relative performance for each depth.

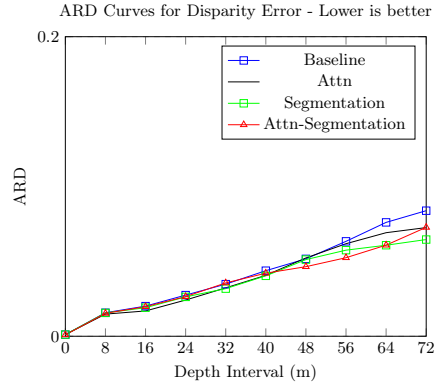
\begin{figure}
\centering
\begin{tikzpicture}[scale=0.7] 
\begin{axis}[
    title={ARD Curves for Disparity Error - Lower is better},
    xlabel={Depth Interval (m)},
    ylabel={ARD},
    xmin=0, xmax=72,
    ymin=0, ymax=0.2,
    xtick={0,8,16,24,32,40,48,56,64,72,80},
    ytick={0,0.2,0.4,0.6,0.8,1.0},
    legend pos=north east,
    ymajorgrids=true,
    grid style=dashed,
]

\addplot[
    color=blue,
    mark=square,
    ]
    coordinates {
    (0, 0.0010826123179867864)(8, 0.015712430700659752)(16, 0.020114369690418243)(24, 0.027477707713842392)(32, 0.0347965732216835)(40, 0.043774448335170746)(48, 0.05167768523097038)(56, 0.0633673146367073)(64, 0.07601942121982574)(72, 0.0838310644030571)
    };

    \addlegendentry{Baseline}
\addplot[color=black]
coordinates {
(0, 0.001115371473133564)(8, 0.014788285829126835)(16, 0.016894491389393806)(24, 0.024178171530365944)(32, 0.03231704607605934)(40, 0.04066726565361023)(48, 0.052282340824604034)(56, 0.06178275868296623)(64, 0.0691361129283905)(72, 0.07255169749259949)
};
\addlegendentry{Attn}
\addplot[
    color=green,
    mark=square,
    ]
        coordinates {
    (0, 0.0011137661058455706)(8, 0.015582454390823841)(16, 0.019075004383921623)(24, 0.026278307661414146)(32, 0.031791411340236664)(40, 0.04033472388982773)(48, 0.051185332238674164)(56, 0.05747772380709648)(64, 0.06077040359377861)(72, 0.06456182152032852) 
    };

    \addlegendentry{Segmentation}

\addplot[
    color=red,
    mark=triangle,
    ]
    coordinates {
    (0, 0.0009489844087511301)(8, 0.015308934263885021)(16, 0.019767917692661285)(24, 0.026446538046002388)(32, 0.0356176421046257)(40, 0.04210490733385086)(48, 0.04645168408751488)(56, 0.05243973806500435)(64, 0.060831792652606964)(72, 0.07272429764270782) 
    };
    \addlegendentry{Attn-Segmentation}

\end{axis}
\end{tikzpicture}
\caption{Comparison of Absolute Relative Difference (ARD) curves showing disparity errors across depth intervals for the three models}
\label{ARD Curve}
\end{figure}

Looking at the ARD curve (Fig. \ref{ARD Curve}) on all model variants we see better performance on the nearer objects with the farther objects having less accuracy. We also note that the self-attention and segmentation mechanisms are having an effect on the mid to distant features and not so much on the nearby objects.

\subsection{Qualitative Analysis}

Based on the predicted disparity images (see Figure \ref{fig:Predicted vs Actual}), we can see that the output is comparable to other methods:

\begin{figure}[h]
    \centering
    \begin{subfigure}{0.3\textwidth}
        \centering
        \includegraphics[width=\textwidth]{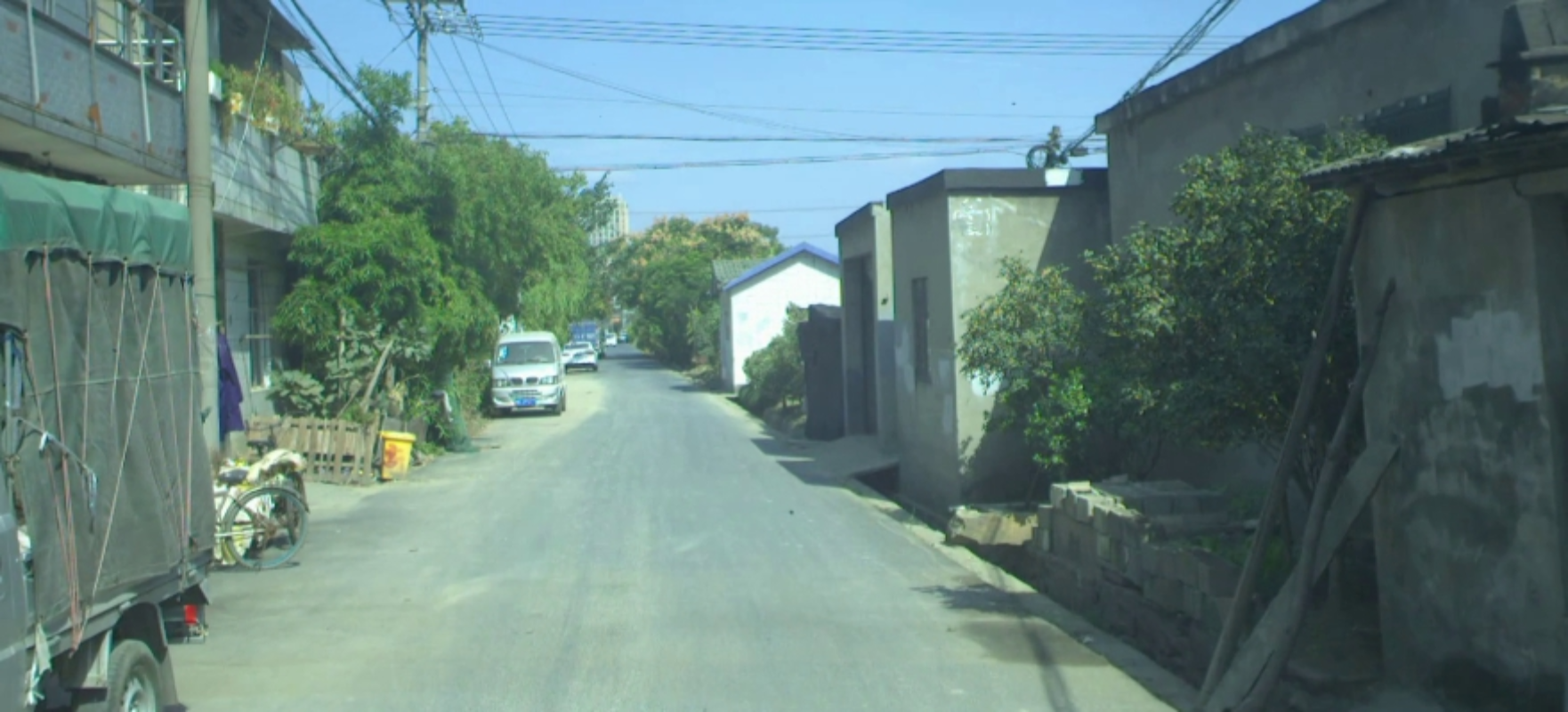}
        \caption{Original Left Image}
    \end{subfigure}
    \begin{subfigure}{0.3\textwidth}
        \centering
        \includegraphics[width=\textwidth]{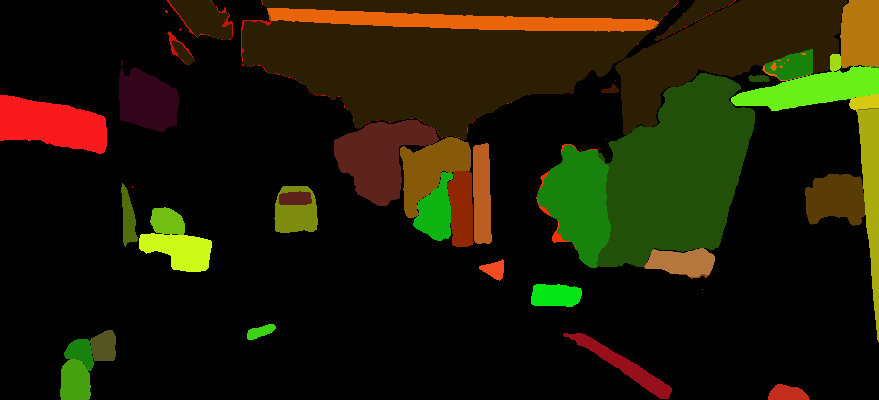}
        \caption{Seg Input via SAM}
    \end{subfigure}
    
    \begin{subfigure}{0.3\textwidth}
        \centering
        \includegraphics[width=\textwidth]{seg_stereo2_disparity.png}
        \caption{USAM-Net output}
    \end{subfigure}
    \begin{subfigure}{0.3\textwidth}
        \centering
        \includegraphics[width=\textwidth]{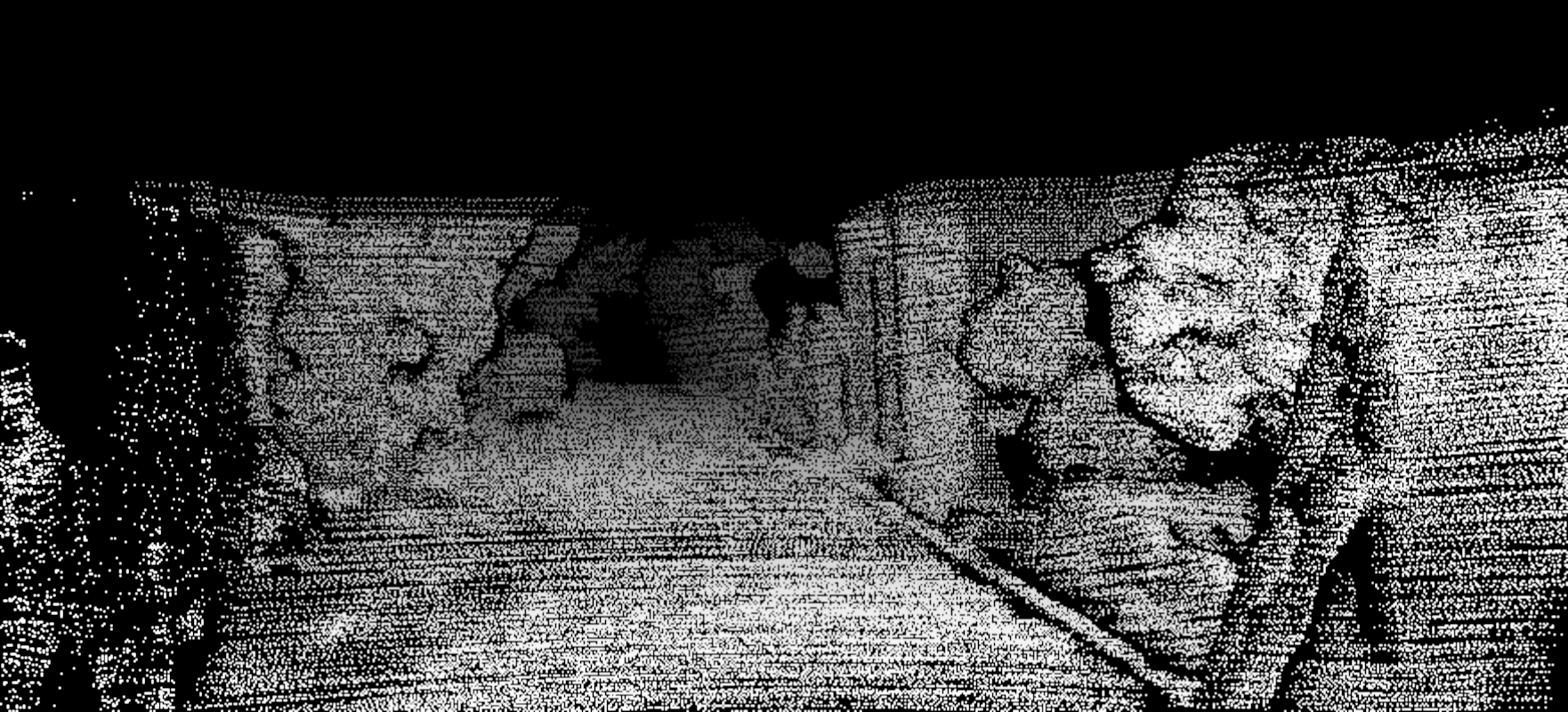}
        \caption{ground truth LIDAR}
    \end{subfigure}
    
    \caption{Predicted vs Actual Disparity Maps}
    \label{fig:Predicted vs Actual}
\end{figure}

\begin{figure}[h]
    \centering
    \begin{subfigure}{0.3\textwidth}
        \centering
        \includegraphics[width=\textwidth]{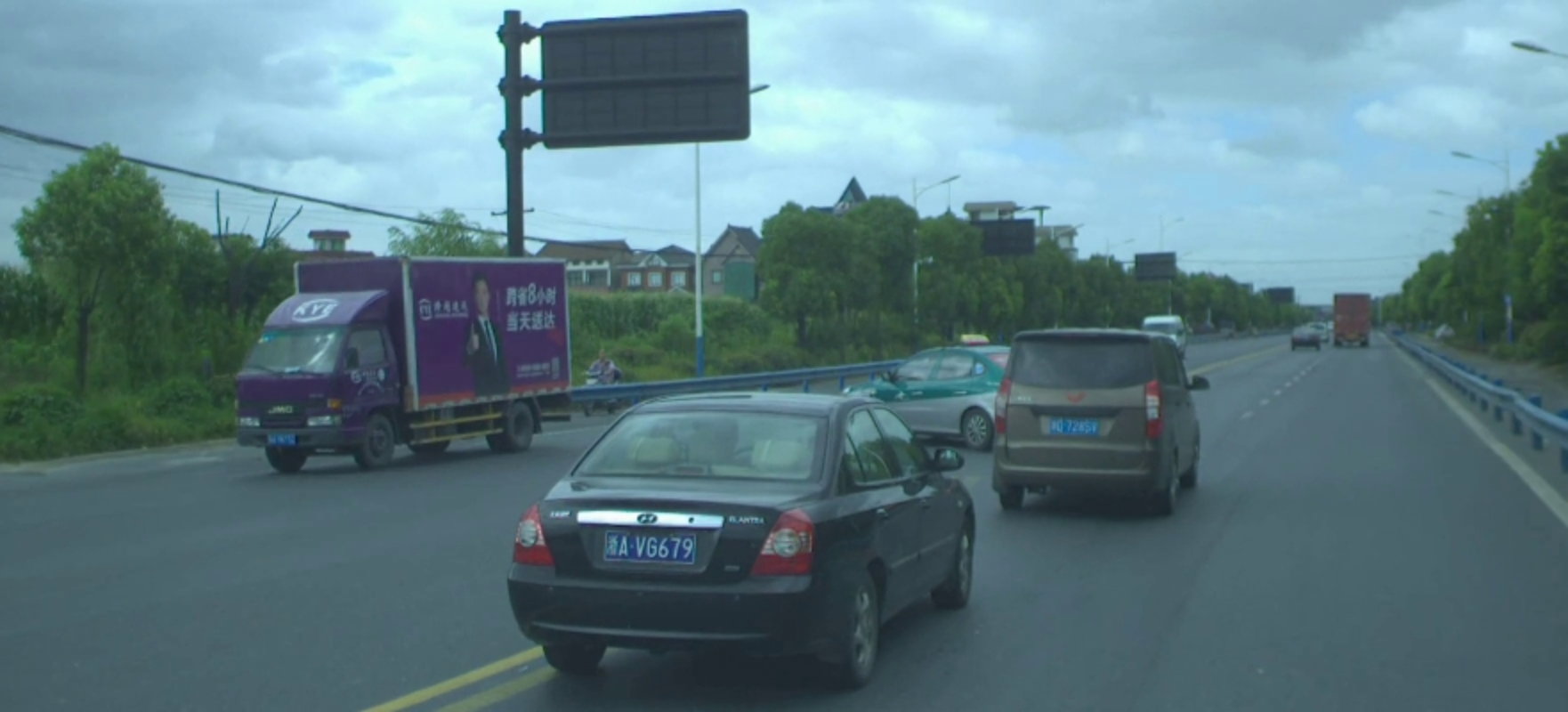}
        \caption{Original Left Image}
    \end{subfigure}
    \begin{subfigure}{0.3\textwidth}
        \centering
        \includegraphics[width=\textwidth]{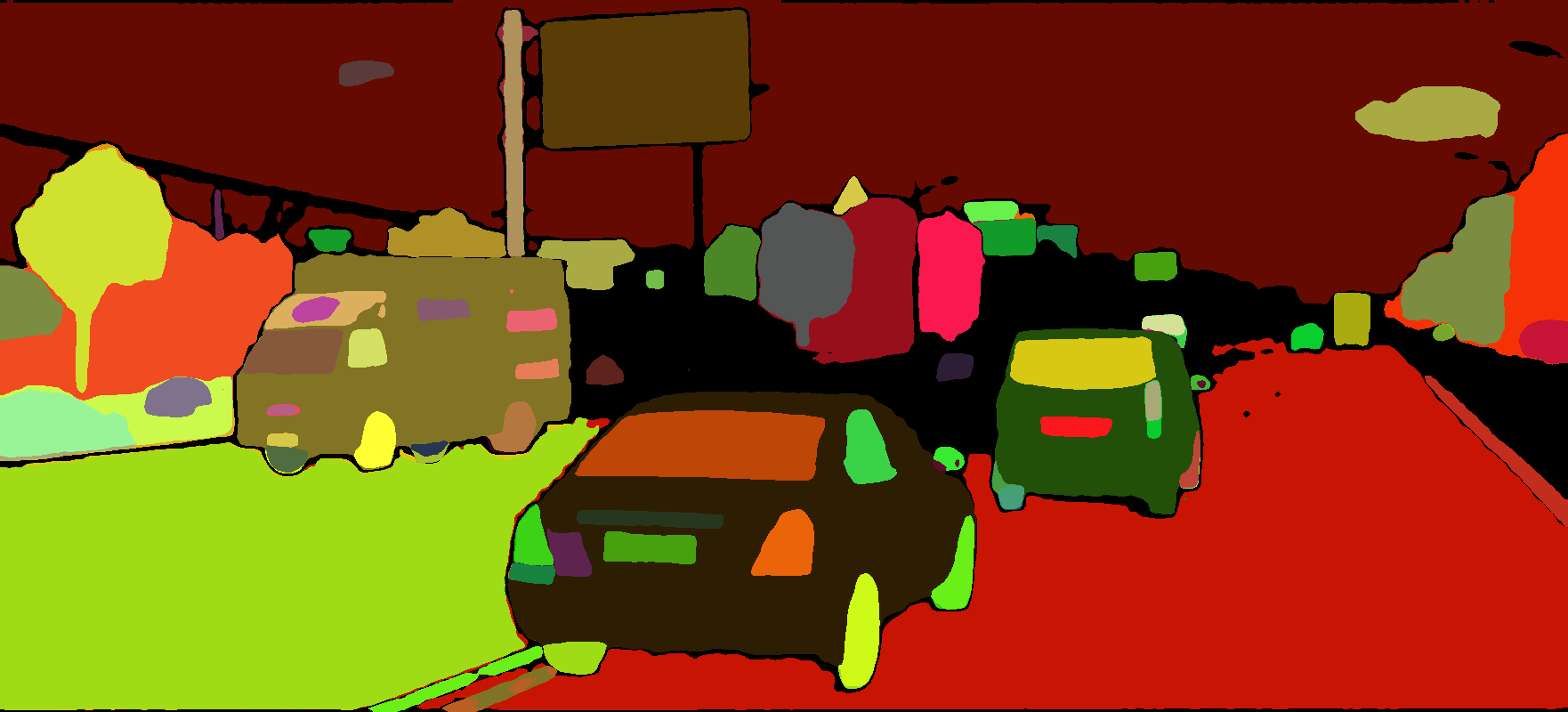}
        \caption{Seg Input via SAM}
    \end{subfigure}
    
    \begin{subfigure}{0.3\textwidth}
        \centering
        \includegraphics[width=\textwidth]{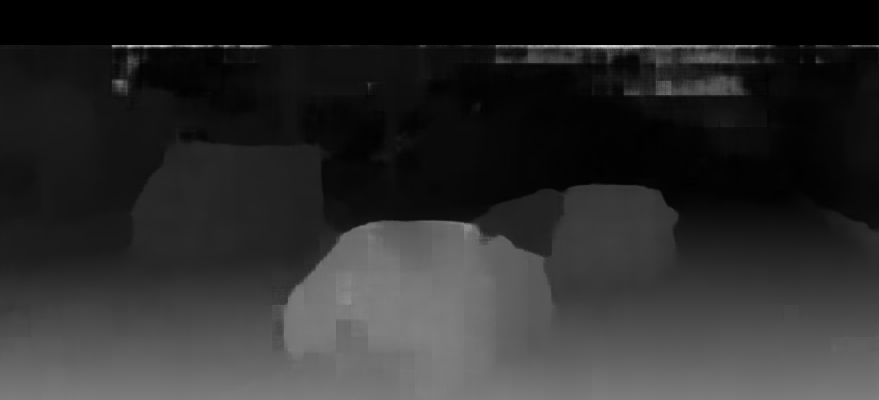}
        \caption{USAM-Net output}
    \end{subfigure}
    \begin{subfigure}{0.3\textwidth}
        \centering
        \includegraphics[width=\textwidth]{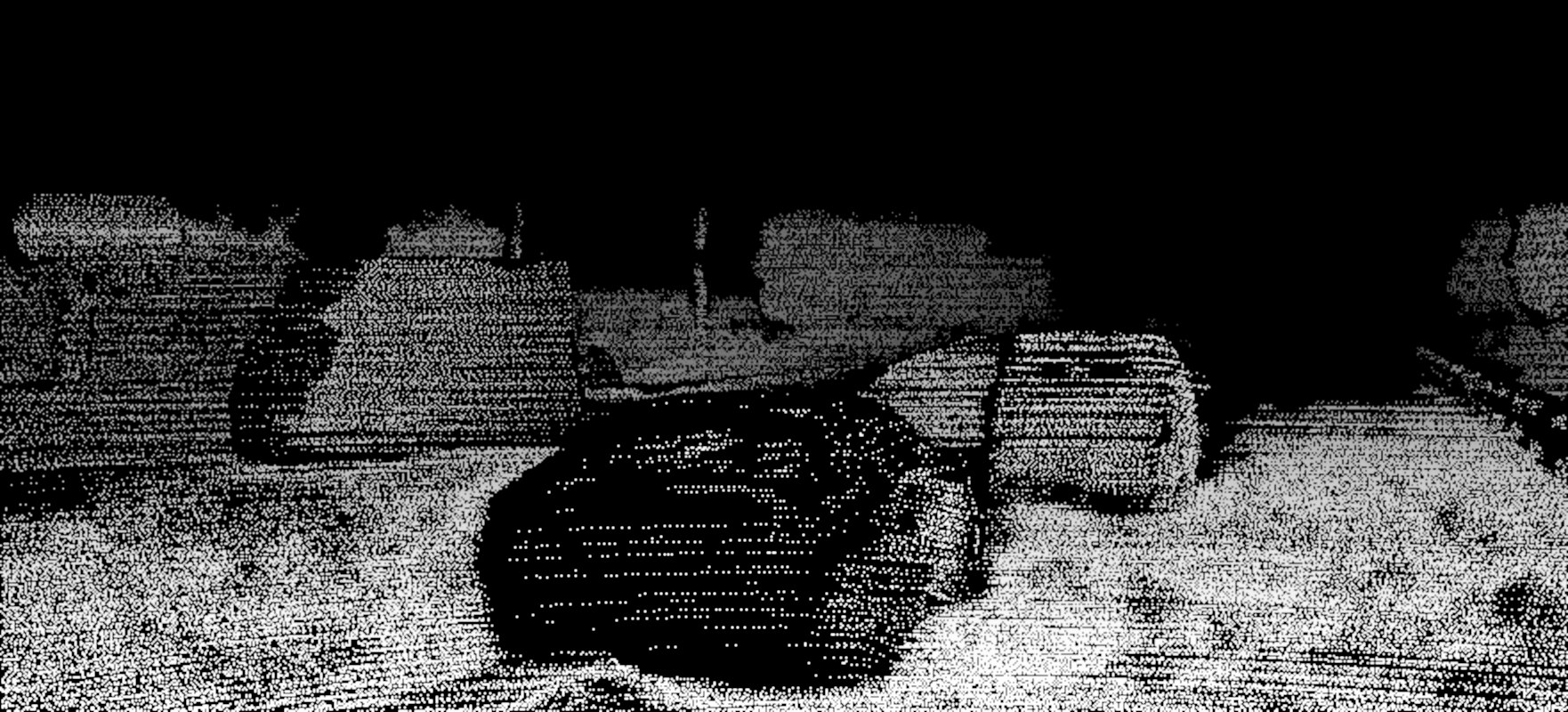}
        \caption{ground truth LIDAR}
    \end{subfigure}
    
    \caption{Predicted vs Actual Disparity Maps}
    \label{fig:Predicted vs Actual 2}
\end{figure}
\begin{figure}[h]
    \centering
    \includegraphics[width=0.3\textwidth]{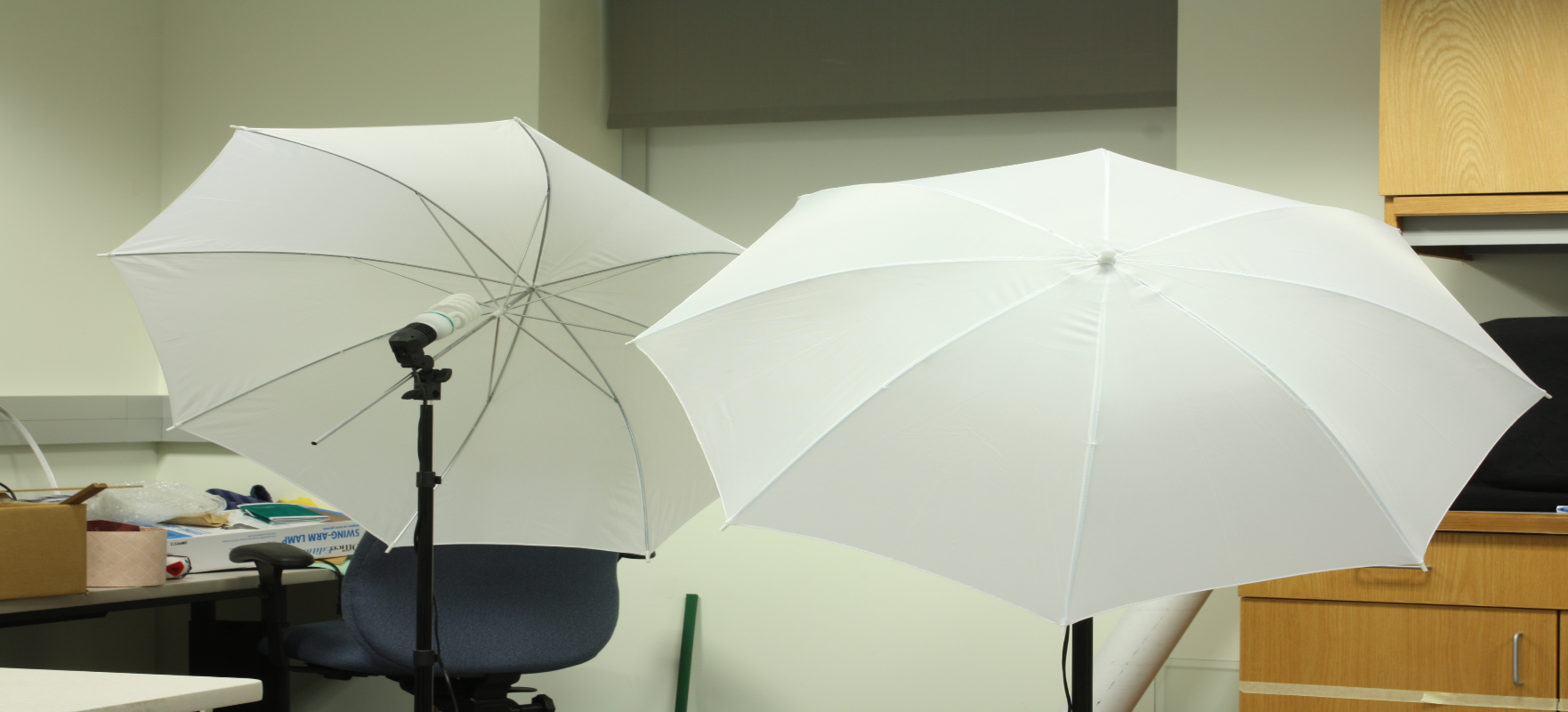}
    \includegraphics[width=0.3\textwidth]{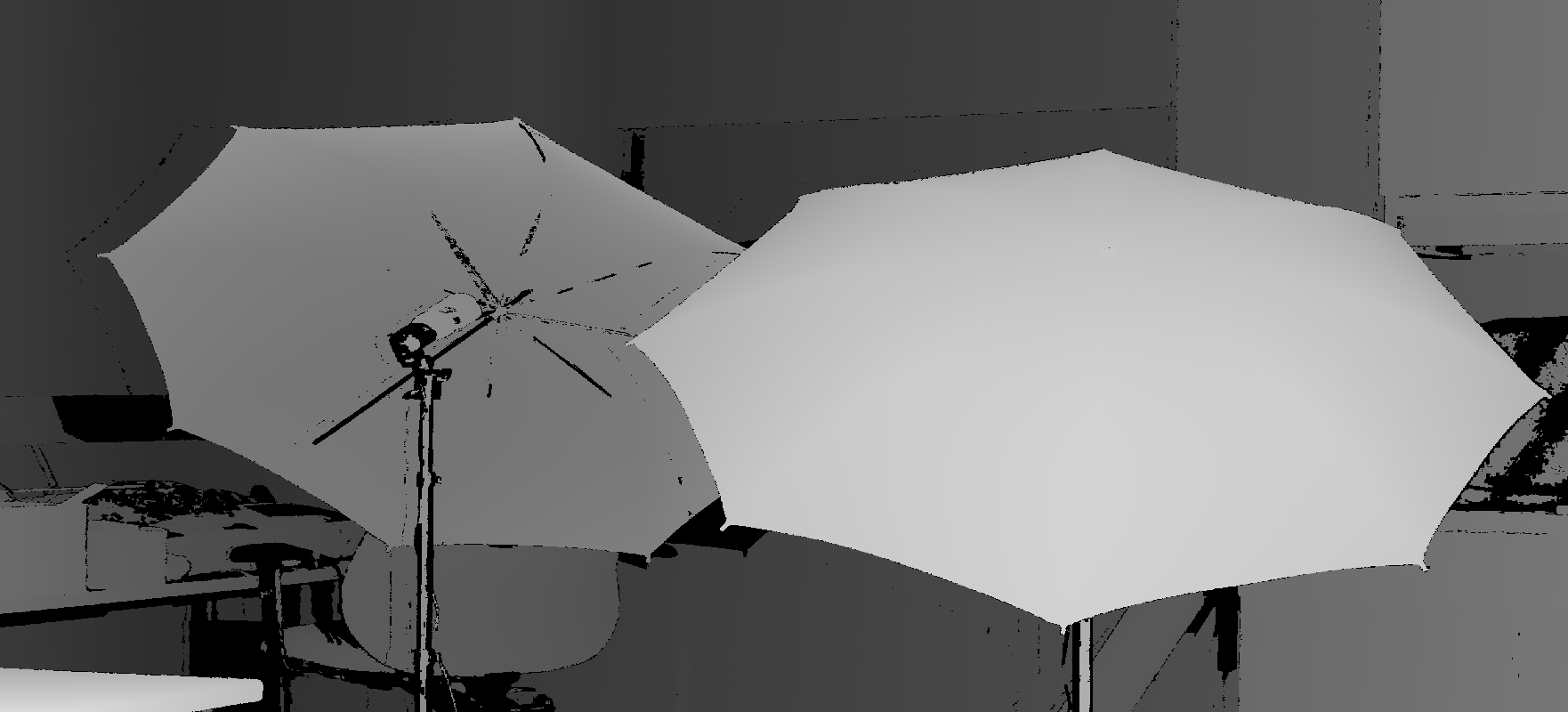}
    \includegraphics[width=0.3\textwidth]{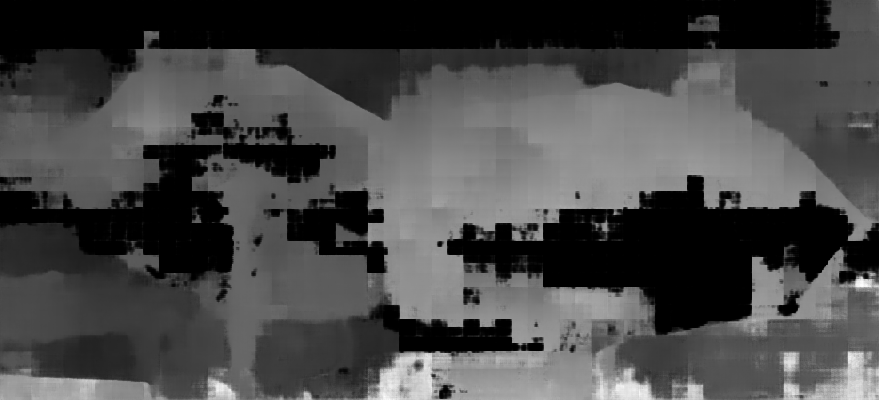}
    \includegraphics[width=0.3\textwidth]{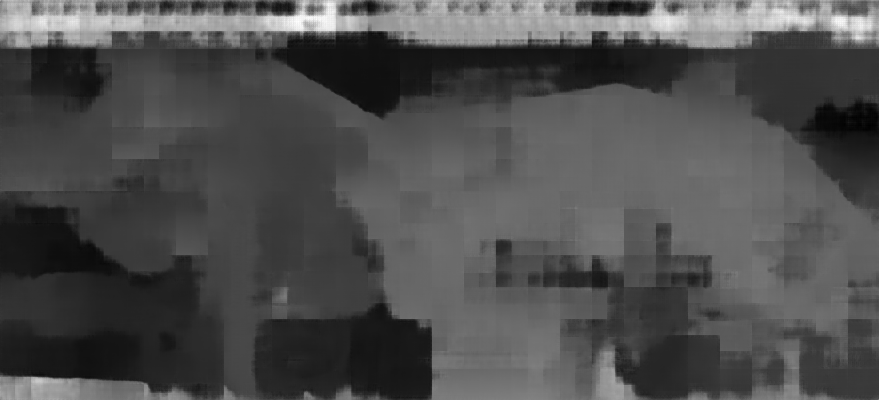}
    \includegraphics[width=0.3\textwidth]{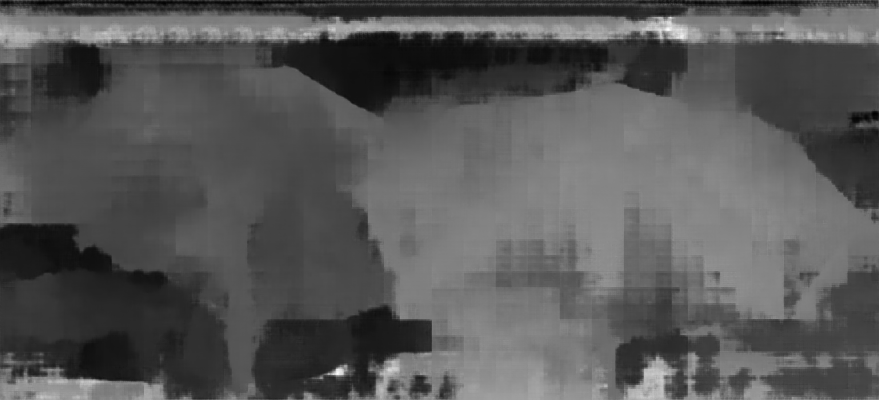}
    \caption{From left to right - Original, Ground Truth, Baseline vs Segmentation vs Attention-Segmentation}
    \label{Predicted vs Actual 2}
\end{figure}

In Figure \ref{Predicted vs Actual 2} we can also see that featureless areas such as roads and flat surfaces are represented correctly in the disparity map. We also note the artifacts in the sky portion of the image that we attempted to rectify in Section \ref{sec:data_preprocess}.

\subsubsection{Performance on KITTI and Middleburry Datasets}

We evaluated the performance of the pre-trained model based on DrivingStereo to the KITTI 2015 evaluation dataset with and without fine-tuning.

\begin{table}[htbp]
\small{
\centering
\caption{KITTI2015 Dataset Benchmarks}
\label{tab:kitti2015_benchmarks}
\begin{tabular}{lcccccccccc}
\toprule
\multirow{2}{*}{Method} & \multicolumn{5}{c}{Before Fine-Tuning} & \multicolumn{5}{c}{After Fine-Tuning} \\
\cmidrule(lr){2-6} \cmidrule(lr){7-11}
 & d1\_all & epe & thres\_1 & thres\_2 & thres\_3 & d1\_all & epe & thres\_1 & thres\_2 & thres\_3 \\
\midrule
Baseline & 8.57 & 1.62 & 46.07 & 16.35 & 8.81 & 5.72 & 1.21 & 30.32 & 10.73 & 5.94 \\
Attn & 7.03 & 1.30 & 34.90 & 13.17 & 7.18 & 5.60 & \textbf{1.11} & 30.83 & 10.81 & 5.79 \\
Seg & 8.27 & 1.46 & 38.97 & 15.42 & 8.52 & 6.10 & 1.27 & 32.08 & 11.80 & 6.35 \\
Segn+Attn & 10.04 & 1.63 & 43.45 & 17.73 & 10.23 & 6.00 & 1.21 & 33.95 & 11.79 & 6.16 \\
\bottomrule
\end{tabular}
\label{kitti2015bench}
}
\end{table}

Table \ref{kitti2015bench} shows that incorporating the attention mechanism significantly improves the model's performance, whereas adding segmentation information has little effect. Although the DrivingStereo and KITTI datasets appear similar, certain characteristics of the KITTI dataset present more challenges to the USAM models.

Also in this evaluation, we test the trained model on the KITTI2015~\cite{Menze2015ISA} and Middlebury~\cite{middlebury} datasets without any fine-tuning. We observe that the model performs well on the KITTI dataset but not as effectively on the Middlebury dataset. This outcome was expected since the objects in the Middlebury dataset differ considerably from those in driving scenarios. Fine-tuning the model may address this issue.

\begin{figure}[h]
\centering
\includegraphics[width=1.0\textwidth]{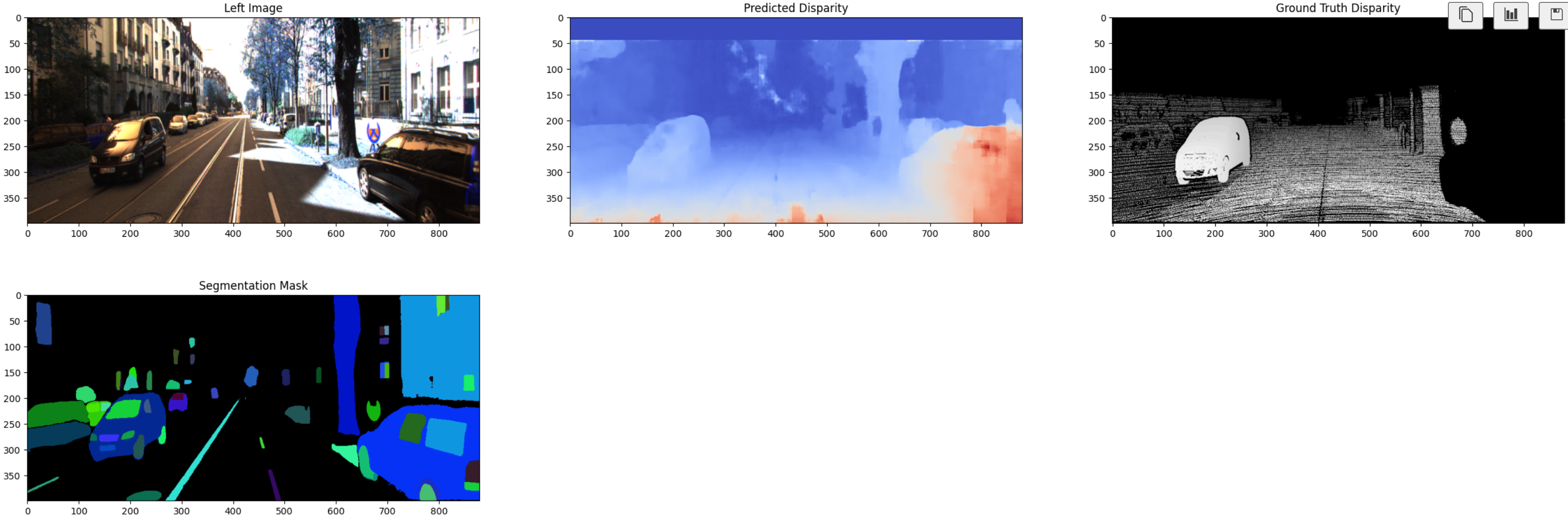}
\caption{Sample Output on the KITTI2015 Dataset}
\label{Output on the KITTI2015 dataset}
\end{figure}

\begin{figure}[ht]
    \centering
    \begin{subfigure}[t]{0.35\textwidth}
        \centering
        \includegraphics[width=0.9\textwidth]{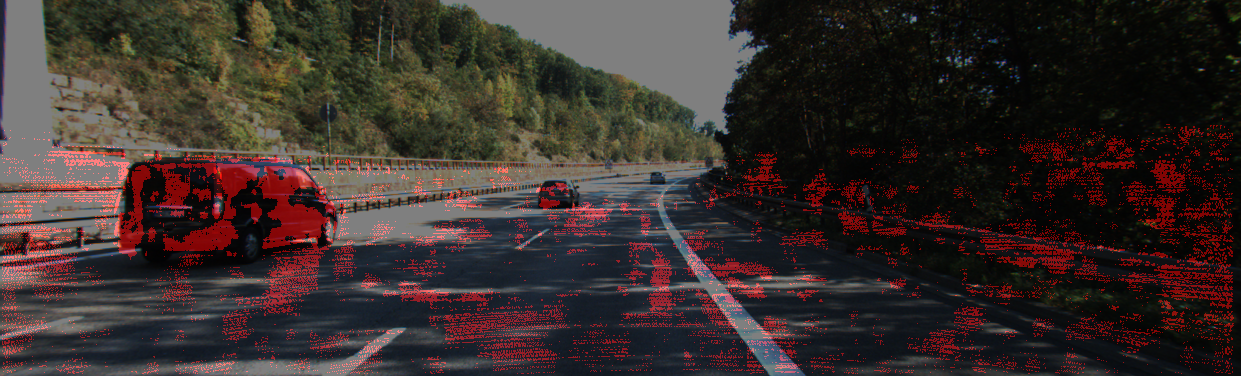}
        \caption{Self-Attention heatmap}
    \end{subfigure}
    \hfill
    \begin{subfigure}[t]{0.35\textwidth}
        \centering
        \includegraphics[width=0.9\textwidth]{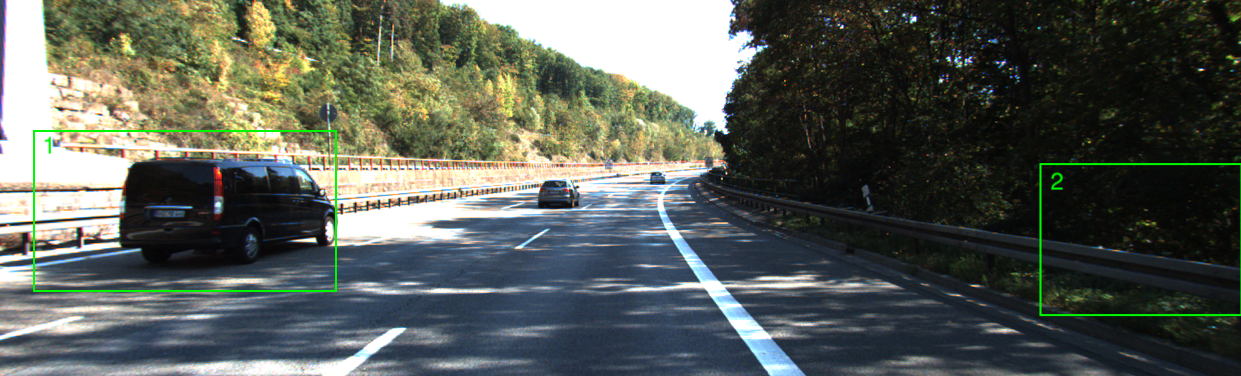}
        \caption{Observed Disparity Regions}
    \end{subfigure}

    \begin{subfigure}[t]{0.25\textwidth}
        \centering
        \includegraphics[width=0.9\textwidth]{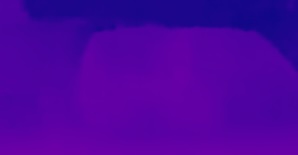}
        \caption{Region 1 - no attn}
    \end{subfigure}
    \hfill
    \begin{subfigure}[t]{0.25\textwidth}
        \centering
        \includegraphics[width=0.9\textwidth]{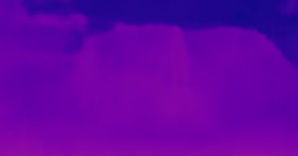}
        \caption{Region 1 - with attn}
    \end{subfigure}

    \begin{subfigure}[t]{0.25\textwidth}
        \centering
        \includegraphics[width=0.9\textwidth]{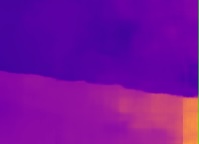}
        \caption{Region 2 - no attn}
    \end{subfigure}
    \hfill
    \begin{subfigure}[t]{0.25\textwidth}
        \centering
        \includegraphics[width=0.9\textwidth]{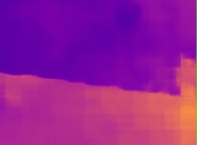}
        \caption{Region 2 - with attn}
    \end{subfigure}

    \caption{Attention Heatmap and Region closeup}
    \label{fig:attention_comparison}
\end{figure}

\subsubsection{Analysis of Self-Attention on the disparity map}

To analyze the effect of the self-attention mechanism with regard to USAM-Net, we compare the variants with and without the self-attention layer on the KITTI 2015 dataset. We compute for difference in disparity between the two variants on the same image and note the regions with the largest difference. We then apply that information as a heatmap overlay on top of the original image.

Based on Fig. \ref{fig:attention_comparison} we can see that the attention mechanism plays a role in areas with many distinctive features. In this case, we can see that the heat map concentrates itself on the car and to a lesser extent on the nearby greenery. The corresponding disparity maps also show increased detail in these areas.

\section{Conclusion and Recommendations}
For this study we have proposed a CNN-based U-Net architecture that uses pre-trained SAM segmentation images and with a self-attention layer. We have demonstrated that this method can successfully learn the representation of the disparity. In addition, we obtained the best GD and EPE scores relative to other models on the DrivingStereo dataset. We have also shown that the attention mechanism works in distinctive feature areas and assists in increasing the detail of the disparity map. We believe that the current USAM network configuration provides a competitive baseline for future work on stereo depth estimation.

Finally, we note that while the quality impact of segmentation on the DrivingStereo and KITTI dataset is small, it can be seen that it has more of an impact on unseen datasets, especially on the Middleburry images.

For further studies and in order to determine the general applicability of the model, it is recommended to run the same model on a different dataset like Sceneflow \cite{MIFDB16} as well as the Middleburry dataset. Although the attention mechanism does not seem to have a big effect on the results that already use segmentation, it has been shown to help in models where segmentation is not provided and is a viable alternative on use cases where running a segmentation network is computationally expensive. Overall, we believe USAM-Net to be a competitive model for highly accurate disparity representation in driving scenarios.

\begin{credits}
{\bf Acknowledgment}
The authors are grateful to the University of the Philippines College of Engineering Artificial Intelligence Program for allowing them to use the NVIDIA DGX A100 HPC System.

\end{credits}

\bibliographystyle{splncs04}
\bibliography{0953}

\begin{thebibliography}{10}
\providecommand{\url}[1]{\texttt{#1}}
\providecommand{\urlprefix}{URL }
\providecommand{\doi}[1]{https://doi.org/#1}

\bibitem{cantrell2020pracdepth}
Cantrell, K.J., Miller, C.D., Morato, C.W.: Practical depth estimation with
  image segmentation and serial u-nets. In: 6th International Conference on
  Vehicle Technology and Intelligent Transport Systems. pp. 406--414 (2020)

\bibitem{githubGitHubJedldOpenStereo}
Dayo, J.E.: {G}it{H}ub - jedld/{O}pen{S}tereo: Fork of {O}pen{S}tereo: {A}
  {C}omprehensive {B}enchmark for {S}tereo {M}atching and {S}trong {B}aseline
  --- github.com. \url{https://github.com/jedld/OpenStereo}, [Accessed
  29-09-2024]

\bibitem{guo2023openstereo}
Guo, X., Lu, J., Zhang, C., Wang, Y., Duan, Y., Yang, T., Zhu, Z., Chen, L.:
  Openstereo: A comprehensive benchmark for stereo matching and strong
  baseline. arXiv preprint arXiv:2312.00343  (2023)

\bibitem{huang2022sdepth}
Huang, B., Zheng, J.Q., Giannarou, S., Elson, D.: H-net: Unsupervised
  attention-based stereo depth estimation leveraging epipolar geometry (2022)

\bibitem{abdullah2023monodepth}
Jan, A., Seo, S.: Monocular depth estimation using res-unet with an attention
  model  (2023)

\bibitem{kirillov2023segany}
Kirillov, A., Mintun, E., Ravi, N., Mao, H., Rolland, C., Gustafson, L., Xiao,
  T., Whitehead, S., Berg, A.C., Lo, W.Y., Doll{\'a}r, P., Girshick, R.:
  Segment anything. arXiv:2304.02643  (2023)

\bibitem{iresnet}
Liang, Z., Feng, Y., Guo, Y., Liu, H., Chen, W., Qiao, L., Zhou, L., Zhang, J.:
  Learning for disparity estimation through feature constancy. In: 2018
  IEEE/CVF Conference on Computer Vision and Pattern Recognition. pp.
  2811--2820 (2018). \doi{10.1109/CVPR.2018.00297}

\bibitem{MIFDB16}
Mayer, N., Ilg, E., H{\"a}usser, P., Fischer, P., Cremers, D., Dosovitskiy, A.,
  Brox, T.: A large dataset to train convolutional networks for disparity,
  optical flow, and scene flow estimation. In: IEEE International Conference on
  Computer Vision and Pattern Recognition (CVPR) (2016),
  \url{http://lmb.informatik.uni-freiburg.de/Publications/2016/MIFDB16},
  arXiv:1512.02134

\bibitem{Menze2015CVPR}
Menze, M., Geiger, A.: Object scene flow for autonomous vehicles. In:
  Conference on Computer Vision and Pattern Recognition (CVPR) (2015)

\bibitem{Menze2015ISA}
Menze, M., Heipke, C., Geiger, A.: Joint 3d estimation of vehicles and scene
  flow. In: ISPRS Workshop on Image Sequence Analysis (ISA) (2015)

\bibitem{Silberman:ECCV12}
Nathan~Silberman, Derek~Hoiem, P.K., Fergus, R.: Indoor segmentation and
  support inference from rgbd images. In: ECCV (2012)

\bibitem{attentionunet}
Oktay, O., Schlemper, J., Folgoc, L., Lee, M., Heinrich, M., Misawa, K., Mori,
  K., McDonagh, S., Hammerla, N., Kainz, B., Glocker, B., Rueckert, D.:
  Attention u-net: Learning where to look for the pancreas  (04 2018).
  \doi{10.48550/arXiv.1804.03999}

\bibitem{unetpaper}
Ronneberger, O., Fischer, P., Brox, T.: U-net: Convolutional networks for
  biomedical image segmentation. vol.~9351, pp. 234--241 (10 2015).
  \doi{10.1007/978-3-319-24574-4_28}

\bibitem{middlebury}
Scharstein, D., Hirschmüller, H., Kitajima, Y., Krathwohl, G., Nešić, N.,
  Wang, X., Westling, P.: High-resolution stereo datasets with
  subpixel-accurate ground truth. vol.~8753, pp. 31--42 (09 2014).
  \doi{10.1007/978-3-319-11752-2_3}

\bibitem{cfnet}
Shen, Z., Dai, Y., Rao, Z.: Cfnet: Cascade and fused cost volume for robust
  stereo matching  (04 2021)

\bibitem{songxiaoedgestereo}
Song, X., Zhao, X., Fang, L., Hu, H., Yu, Y.: Edgestereo: An effective
  multi-task learning network for stereo matching and edge detection.
  International Journal of Computer Vision  \textbf{128} (04 2020).
  \doi{10.1007/s11263-019-01287-w}

\bibitem{victor2021survey}
Victor, V.S., Neigel, P.: Survey on semantic stereo matching/semantic depth
  estimation. arXiv preprint arXiv:2109.10123  (2021)

\bibitem{xu2023iterativegeometryencodingvolume}
Xu, G., Wang, X., Ding, X., Yang, X.: Iterative geometry encoding volume for
  stereo matching (2023), \url{https://arxiv.org/abs/2303.06615}

\bibitem{yang2019drivingstereo}
Yang, G., Song, X., Huang, C., Deng, Z., Shi, J., Zhou, B.: Drivingstereo: A
  large-scale dataset for stereo matching in autonomous driving scenarios. In:
  IEEE Conference on Computer Vision and Pattern Recognition (CVPR) (2019)

\bibitem{yang2018segstereo}
Yang, G., Zhao, H., Shi, J., Deng, Z., Jia, J.: Segstereo: Exploiting semantic
  information for disparity estimation (2018)

\end{thebibliography}

%
%
%
%
\end{document}